\PassOptionsToPackage{numbers,sort&compress}{natbib}
\documentclass{ijisa}

\usepackage[utf8]{inputenc}
\usepackage[english]{babel}
\usepackage{newtxtext,newtxmath}
\usepackage{graphicx}
\usepackage{amsmath}  
\usepackage{subcaption}
\usepackage{booktabs}
\usepackage{siunitx}
\usepackage{float}
\usepackage{xcolor}
\usepackage{enumitem}
\usepackage{tikz}
\usetikzlibrary{arrows.meta,positioning,calc,fit}
\definecolor{accentgold}{HTML}{D9A441}
\usepackage{hyperref}
\hypersetup{
  colorlinks = true,
  urlcolor   = blue,
  linkcolor  = blue,
  citecolor  = blue
}
\setlist{nolistsep,leftmargin=*}
\graphicspath{{figures/}}

\title{Risk-Sensitive Mission Planning for Multi-Target On-Orbit Inspection via Distributional Reinforcement Learning\\}

\author[1,*]{Adam Abdin}

\affil[1]{Universit\'{e} Paris-Saclay, CentraleSup\'{e}lec, Department of Industrial Engineering, Gif-sur-Yvette, France.\vspace*{.2cm}}
\affil[*]{Corresponding author. e-mail: adam.abdin@centralesupelec.fr}

\abstract{
Multi-target on-orbit servicing missions require planning tools that can adapt to evolving operational conditions while accounting for the outcome variability inherent to safety-critical spaceflight. This paper presents an adaptive planning approach for multi-target servicing campaigns in low Earth orbit and applies it to inspection missions flown by a low-thrust servicer. The approach is based on Distributional Reinforcement Learning, which learns the full distribution of mission returns and thereby supports explicit risk-sensitive decision-making at each planning step. Specifically, a Quantile-Regression Deep Q-Network (QR-DQN) agent selects which target to inspect next by jointly considering the expectation and the variance of the predicted return through a mean--variance criterion. The mission environment models circular orbits with a spread in inclination and uses Edelbaum's combined-manoeuvre low-thrust transfer estimates. The observation comprises the orbital elements, the visit history, the observed risk state of each target and the remaining mission-time budget. The risk state of each target changes stochastically during the campaign, making the outcome of an inspection uncertain and creating operationally meaningful distinctions between risk-neutral and risk-averse planning strategies. Because the mission-time budget generally admits only a subset of the candidate targets, the planner must also select which targets to inspect. Numerical experiments on scenarios with five target satellites show that one risk-aversion weight moves the policy across a range of mean return and return variance, over an unchanged environment and reward. Risk-averse agents reduce the variance of the episodic return by approximately one half relative to risk-neutral agents and lower the probability of a mission loss, at a cost in mean return. To validate the learned policies, the efficient frontier, which gives the highest mean return attainable at each level of return variance, is computed exactly by enumerating the admissible visit orders. At equal variance, the learned policies lie close to this frontier in mean return, and the risk-averse policies have a lower probability of a mission loss than the frontier policies.
}


\begin{document}

\maketitle
\abstract
\keywords
\thispagestyle{thefirstpage}

\section*{Nomenclature}
\begin{description}[leftmargin=5.4em,labelwidth=4.8em,labelsep=0.6em,font=\normalfont]
\item[LEO] Low Earth Orbit
\item[MDP] Markov Decision Process
\item[OOS] On-Orbit Servicing
\item[QR-DQN] Quantile-Regression Deep Q-Network
\item[RL] Reinforcement Learning
\item[$a,i$] Semi-major axis, inclination
\item[$a_T$] Thrust acceleration
\item[$B$] Mission time budget
\item[$b$] Residual mission-time budget
\item[$F,m$] Thrust, servicer mass
\item[$K$] Return quantiles learned per action
\item[$L$] High-risk failure loss
\item[$N$] Candidate targets
\item[$n_e$] Targets in the high-risk state
\item[$p$] Failure probability of a high-risk inspection
\item[$p_j(t)$] Failure probability of the inspection of target $j$ at time $t$
\item[$q_{0.1}$] Lower decile of the mission return
\item[$r_t$] One-step reward (inspection outcome)
\item[$v_j$] Inspection value of target $j$
\item[$Z$] Mission return (sum of inspection rewards)
\item[$z_k$] Learned quantile estimates of $Z$
\item[$\alpha$] Action (target selected for inspection)
\item[$\beta$] High-risk success premium
\item[$\gamma$] Discount factor (training only)
\item[$\lambda$] Mean--variance risk-aversion weight
\item[$\theta$] Plane change between two orbits
\item[$\mu$] Earth's gravitational parameter
\item[$\rho$] Risk-adjusted action score
\item[$\Delta t$] Transfer duration
\item[$\Delta v$] Velocity increment (manoeuvre cost)
\end{description}

\section{Introduction}
\label{section:intro}

On-orbit servicing (OOS), encompassing inspection, maintenance, refuelling, and removal of spacecraft already deployed in orbit, remains a central focus of ongoing research efforts. This sustained interest is driven by the significance of extending the operational lifespan of space assets and the escalating congestion within the orbital environment \citep{sorenson_multi-orbit_2023,apa_optimal_2025-1}. Deploying a dedicated spacecraft for each asset requiring servicing is uneconomical. As a result, a servicing spacecraft is anticipated to visit and service multiple targets during a single campaign. The task of selecting targets and establishing a visit sequence, while considering constraints on fuel, time, and operations, is a complex planning challenge. This challenge must be navigated under conditions of uncertainty, as the demand for servicing and the outcome of each servicing operation are unknown at the planning stage.

Such a campaign has the structure of an orienteering problem \citep{vansteenwegen_orienteering_2011}, wherein a subset of candidate targets is selected and arranged within a predetermined travel budget. Visiting each target necessitates expenditure of time and fuel, which are determined by the transfer between orbits. The value of a servicing operation depends on client demand, which may fluctuate during the mission, and the outcome of each operation is uncertain. Given a limited mission budget, a servicing spacecraft generally cannot service all potential targets and must therefore make strategic selections regarding which requests to fulfil.

A high-priority but operationally degraded request may, on average, be worth servicing, although a failed operation incurs a significant loss to the mission. A planner that optimises solely for the expected return treats such a request identically to one of the same average value that cannot fail, despite differences in their return distributions in terms of spread and lower tail.

Each servicing operation decreases the remaining budget for future targets, so the value of servicing a particular target depends on which others are still reachable. Sequential decisions of this kind are formulated as Markov decision processes (MDPs), and reinforcement learning (RL) in particular estimates this value from experience across the scenario distribution \citep{sutton_reinforcement_2020}. Traditional RL focuses on maximising the expected return, making it risk-neutral. In contrast, distributional RL models the probability distribution of the return \citep{bellemare_distributional_2017,bellemare_distributional_2023}, thereby enabling the use of risk measures at decision time. An MDP in which policies are ranked by a risk measure of the return is a risk-sensitive MDP \citep{mannor2011meanvariance}. To the best of our knowledge, distributional risk-sensitive RL has not been applied to multi-target OOS mission planning.

This paper quantifies the trade-off between expected return and exposure to failed operations in a servicing campaign, examining whether this balance can be controlled by the planner. It formulates multi-target servicing planning within a mission-time budget and under a stochastic risk of failure at each target as a risk-sensitive MDP, featuring closed-form low-thrust transfer costs and a masked discrete action space. The problem is solved with a quantile-regression deep Q-network (QR-DQN) that applies a mean--variance criterion to the learned return distribution, where one weight sets both the mean return and the return variance targeted by the policy. The approach is applied to inspection campaigns flown by a low-thrust servicer over five targets in low Earth orbit (LEO), inspired by the Iridium-33 debris cloud. For these campaigns, the efficient frontier of mean return against return variance, which gives the highest mean return attainable at each level of variance, is computed exactly by enumerating the admissible visit sequences. The paper reports the mean return, return variance, and lower tail of the learned policies at each weight setting, assesses the proportion of the frontier these policies recover, and shows how the weight can be set from a required probability of a mission loss.

\section{Related Work}
\label{section:related}

\subsection{Multi-target mission planning and low-thrust transfers}

Planning a campaign that visits multiple orbital targets constitutes a longstanding challenge in astrodynamics, with the cost of each trajectory segment determined by the transfer model. In the context of low-thrust, electric-propulsion regimes pertinent to extended servicing missions, \citet{perkins1959flight} and \citet{edelbaum_propulsion_1961} established foundational approximations relating the gradual spiral between circular orbits to its associated velocity expenditure. Due to the computational expense of integrating continuous-thrust equations of motion for every potential trajectory, rapid analytic approximations have been developed for multi-target scenarios \citep{gatto_fast_2015}.

At the planning level, early formulations treat servicing as a routing problem. \citet{bourjolly_onorbit_2006} model on-orbit servicing as a time-dependent, moving-target travelling salesman problem that minimises the total energy or time of the campaign, and \citet{izzo_evolving_2015} introduce travelling-salesman variants for debris removal in which the subset of debris to visit is selected. \citet{sorenson_multi-orbit_2023} investigate multi-orbit routing and scheduling for servicing spacecraft under the assumption of highly manoeuvrable vehicles, and \citet{apa_optimal_2025-1} determine optimal visit sequences for a low-thrust servicing spacecraft through mixed-integer programming based on analytic transfer-cost estimates. \citet{bandyopadhyay_revisiting_2024} employ masked Proximal Policy Optimization for multi-debris rendezvous sequencing, and report shorter mission durations than genetic algorithms and greedy baselines. This masked-policy framework has since been extended to include refuelling and collision avoidance within the Iridium-33 debris environment \citep{bandyopadhyay_optimizing_2026}.

Uncertainty in servicing campaigns has been addressed in several ways. \citet{sarton_framework_2021} maximise the profit of servicing operations for geostationary satellites with mixed-integer programming, re-solved over a rolling horizon as random service needs arise. \citet{patnala_optimal_2026} formulate the dispatch of a servicing vehicle under stochastic demand and partial observability as a partially observable MDP. \citet{patnala_onorbit_2026} train a servicer with the cross-entropy method to rendezvous with satellites under conjunction threat and perform collision-avoidance manoeuvres, including sequential manoeuvres for several satellites, and \citet{nikitits_risk_2026} formulate active debris removal as an MDP with a dynamic collision risk carried in the state. In these studies, the objective is a deterministic cost or profit, re-optimised as uncertainty is revealed, or an expected cumulative reward. In contrast to these studies, this paper addresses the selection of a subset of targets under a stochastic risk of failure at each target, and a risk measure is applied to the learned return distribution at decision time. The spread of the mission return is thereby set through the decision criterion.

\subsection{Distributional reinforcement learning}

Distributional methods for RL differ in how they represent the return distribution \citep{bellemare_distributional_2023}. The categorical approach, known as C51, assigns the return to a predetermined support \citep{bellemare_distributional_2017}, which constrains accuracy when returns extend beyond the specified range. In QR-DQN, the cumulative probabilities are fixed while the quantile values are learned, thereby overcoming this limitation \citep{dabney_distributional_2018}. Subsequently, other methods learn a continuous quantile function \citep{dabney_implicit_2018}. This study adopts QR-DQN and assesses the mean--variance criterion based on its quantile estimates.

\subsection{Risk-sensitive reinforcement learning}

A risk measure converts a return distribution into a single value that an agent can optimise. The mean--variance criterion, which penalises return variance, originates from portfolio theory \citep{markowitz_1952_portfolio} and has been used in RL through policy-gradient techniques \citep{castro_policy_2012}. The conditional value-at-risk, representing expected return in the worst tail, is common in stochastic optimisation \citep{rockafellar2000cvar} and has been combined with distributional RL for risk-sensitive control \citep{lim_distributional_2022}. In safety-critical spaceflight, a rare failed operation can cost more than many routine returns, so mission planners are concerned with the spread and the lower tail of the mission return as well as its mean. Both criteria can be computed from the same quantile estimates. This paper uses the mean--variance criterion, which penalises the dispersion of the return through a single weight.

\section{Problem Formulation}
\label{section:formulation}

\subsection{Scenario}

\begin{figure}[t]
\centering
\includegraphics[width=0.94\columnwidth]{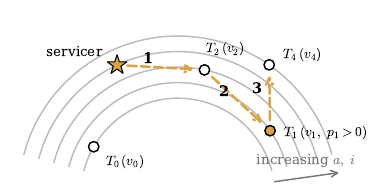}
\caption{Mission concept overview. A low-thrust servicer transfers between circular orbits with varying semi-major axes and inclinations to inspect multiple targets within a set mission-time limit. Each target $j$, marked $T_j$, has a value $v_j$ and might enter a high-risk state with a failure probability $p_j>0$ (indicated by a filled marker). Numbered arrows illustrate a sample visit sequence.}
\label{fig:concept}
\end{figure}

We consider a single low-thrust servicing spacecraft operating among $N$ potential targets in LEO. The missions considered in this paper are inspection campaigns, in which the operator sends the servicer to acquire data on the condition of each target, for example to characterise debris objects or to monitor satellites of interest. Each target $j$, ranging from $1$ to $N$, is situated in a circular orbit characterised by its semi-major axis $a_j$ and inclination $i_j$. All targets are taken to share a common ascending node, so the transfer cost depends only on $a_j$ and $i_j$. The servicing spacecraft begins co-located with one target and is allocated a fixed mission-time budget $B$. Given constant thrust acceleration, the transfer duration is proportional to the change in velocity, indicating that $B$ is a velocity budget and defines a limited operational campaign window. Each target has an inspection value $v_j$, the priority the operator assigns to its inspection. The planner selects the sequence of targets to inspect, and only a subset of the $N$ candidates can typically be covered within the budget. Figure~\ref{fig:concept} illustrates this mission concept.

\subsection{Transfer cost}

The servicer uses a continuous low-thrust engine with thrust $F$ and mass $m$, resulting in a constant acceleration $a_T=F/m$. The transfer cost between two circular orbits is based on the combined manoeuvre of \citet{edelbaum_propulsion_1961}, which accounts for simultaneous changes in altitude and orbital plane. For circular speeds $V=\sqrt{\mu/a}$ and $V_j=\sqrt{\mu/a_j}$, where $\mu$ is Earth's gravitational parameter, and a plane change of $\theta=\lvert i-i_j\rvert$ in radians, the velocity increment is given by
\begin{equation}
\Delta v = \sqrt{V^{2}+V_j^{2}-2\,V V_j\cos\!\big(\tfrac{\pi}{2}\theta\big)}.
\label{eq:edelbaum}
\end{equation}
The transfer time is calculated as $\Delta t = \Delta v / a_T$, using the constant acceleration. In a coplanar transfer ($\theta=0$), Eq.~(\ref{eq:edelbaum}) simplifies to $\Delta v=\lvert V-V_j\rvert$. Equation~(\ref{eq:edelbaum}) provides an analytic estimate of the transfer cost, following the fast low-thrust approximation of \citet{gatto_fast_2015}, allowing the planner to evaluate many potential transfers without full numerical integration.

\subsection{Stochastic inspection risk}

The risk status of the uninspected targets is updated periodically. For each target $j$, the inspection has a failure probability $p_j(t)$, which is either $0$ or $p$. An inspection is considered nominal if $p_j=0$, or high-risk if $p_j=p>0$. High-risk inspections are those performed near a tumbling or non-cooperative target, where an uncertain attitude state or a keep-out violation may cause the inspection to be abandoned. After each redraw, $n_e$ of the uninspected targets, or all of them when fewer remain, are randomly selected without replacement to be in the high-risk state, with probability proportional to their value, under the assumption that higher-value targets are more safety-critical. Inspecting a high-risk target removes it from this set. Inspecting target $j$ results in an outcome:
\begin{equation}
r_j=
\begin{cases}
v_j, & p_j=0,\\[2pt]
v_j+\beta \ \ \text{w.p. } 1-p, \quad -L \ \ \text{w.p. } p, & p_j=p,
\end{cases}
\label{eq:demand}
\end{equation}
a successful high-risk inspection yields a premium $\beta$ over the nominal value, while a failed one results in a loss $L$. A failed inspection still consumes the transfer time, and the target is not revisited. The premium is set so that a high-risk inspection offers a higher expected value and more return variability compared to a nominal inspection. Skipping a high-risk target therefore results in a loss of expected value. A risk-neutral planner prioritises high-value, high-risk targets and accepts occasional significant losses, whereas a risk-averse planner may forgo some expected value to minimise the risk of large losses.

\subsection{Markov decision process}
\label{subsec:mdp}

The campaign is modelled as an MDP characterised by a state space $\mathcal{S}$, an action space $\mathcal{A}$, a transition kernel $\mathcal{P}$, a reward function $\mathcal{R}$, and a discount factor $\gamma$. The selection of a target influences the transfer time, the set of inspected targets, and the remaining budget. The inspection outcome is stochastic, so the mission return $Z$, the sum of the rewards collected over the campaign, is a random variable, and each policy is characterised by the distribution of $Z$ it induces.

\textbf{State:} The observation includes, for both the servicer and all targets, the normalised orbital elements $a$ and $i$, a binary indicator of whether the target has been inspected, and the observed failure probability $p_j(t)$ of each target. Additionally, it comprises the normalised value of each target, the remaining budget, and the valid-action mask, as defined below. For $N$ targets, the observation vector has a length of $6N+3$.

\textbf{Action and Masking:} The action $\alpha$ signifies the next target to inspect. Only targets that remain uninspected and are reachable within the remaining budget are considered admissible, as per Equation~(\ref{eq:mask}),
\begin{equation}
\mathcal{A}_{\mathrm{feas}}=\big\{\,j:\ \text{$j$ uninspected},\ \Delta t_{c\to j}\le b\,\big\},
\label{eq:mask}
\end{equation}
where $\Delta t_{c\to j}$ denotes the transfer time from the current target $c$ to target $j$, and $b$ represents the residual budget. Inadmissible actions are excluded by masking, which assigns their values negative infinity before the greedy selection. The servicer is thus restricted to feasible inspections, aligning with the masked policies described by \citet{bandyopadhyay_revisiting_2024}. The servicer continues inspection activities while any uninspected target remains affordable, and the campaign concludes when the available budget cannot accommodate subsequent inspections. Risk aversion is thus reflected in the choice among the admissible inspections.

\textbf{Reward and objective:} The one-step reward $r_t$ is the inspection outcome (\ref{eq:demand}) of the chosen target, and $Z=\sum_t r_t$ is the collected value. A risk-neutral planner aims to maximise the expected $Z$. Conversely, risk-sensitive planners described in Section~\ref{section:method} choose actions based on a risk-adjusted score derived from the predicted return distribution. During training, a discount factor $\gamma$ is incorporated into the temporal-difference target, although the reported returns are undiscounted.

\section{Distributional Risk-Sensitive Solution}
\label{section:method}

\begin{figure}[t]
\centering
\begin{tikzpicture}[
  font=\scriptsize,
  box/.style={draw=black, line width=0.5pt, rounded corners=2pt, align=center,
              inner sep=4pt, fill=white, text width=5.0cm},
  sig/.style={-{Stealth[length=2mm]}, line width=0.6pt},
  slab/.style={font=\scriptsize, inner sep=1.5pt, fill=white}]
\node[box] (env) {\textbf{Mission environment}\\[1pt]
  orbital transfers and inspection risk};
\node[box, below=0.62cm of env] (net) {\textbf{Quantile network}\\[1pt]
  return quantiles $\{z_k(s_t,\alpha)\}_{k=1}^{K}$ per action};
\node[box, below=0.62cm of net, draw=accentgold, line width=0.9pt] (risk) {%
  \textbf{Risk-sensitive selection}\\[1pt]
  mean--variance score $\rho^{\mathrm{MV}}_{\lambda}$\\
  greedy selection under the feasibility mask};
\draw[sig] (env) -- (net) node[slab,midway]{state $s_t$};
\draw[sig] (net) -- (risk);
\draw[sig] (risk.east) -- ++(0.62,0) |- (env.east)
  node[slab,pos=0.25,xshift=6pt]{action $\alpha_t$};
\end{tikzpicture}
\caption{Solution architecture. The observation $s_t$ is formed in the mission environment. The quantile network (inverse cumulative distribution) returns $K$ quantiles per action, and the next inspection is selected from them using the mean--variance score under the feasibility mask. The same masked score is applied to the bootstrap action during training.}
\label{fig:arch}
\end{figure}
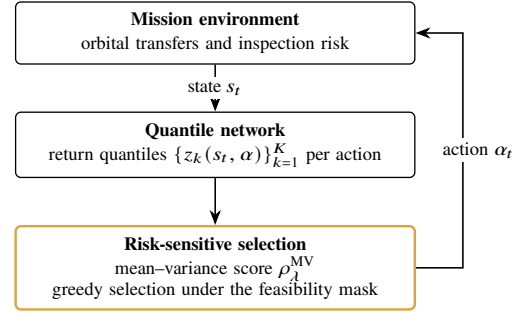

Figure~\ref{fig:arch} provides an overview of the solution architecture, including the mission environment described in Section~\ref{section:formulation}, the quantile network, and the masked risk-sensitive action selection.

\subsection{Learning the return distribution}

Standard value-based RL estimates the expected action value $Q(s,\alpha)$, representing the expected return $Z$ conditioned on state $s$ and action $\alpha$. Using the approach of \citet{dabney_distributional_2018}, the distribution of $Z(s,\alpha)$ is modelled with quantiles. The network provides, for each action, $K$ estimates $z_k(s,\alpha)$, corresponding to quantiles at fixed, evenly spaced cumulative probabilities $\hat\tau_k=(k-\tfrac12)/K$, with $k$ from $1$ to $K$. These quantiles are trained by minimising the quantile-Huber loss against the one-step distributional Bellman target, so that the $z_k$ approximations reflect the inverse cumulative distribution of $Z(s,\alpha)$. The risk-adjusted score of Section~\ref{subsec:mv} is computed directly from these quantile values.

\subsection{Risk-sensitive action selection}
\label{subsec:mv}

The predicted quantiles of $Z(s,\alpha)$ are transformed into a risk-adjusted score $\rho(s,\alpha)$, which the policy uses greedily within the admissible set (\ref{eq:mask}). The mean--variance criterion \citep{markowitz_1952_portfolio,castro_policy_2012} penalises return variance, as defined in Eq.~(\ref{eq:mv}),
\begin{equation}
\rho^{\mathrm{MV}}_\lambda(s,\alpha)=\widehat{\mathbb{E}}[Z]-\lambda\,\widehat{\mathbb{V}}[Z]
=\frac{1}{K}\sum_k z_k-\lambda\,\widehat{\mathbb{V}}\big(\{z_k\}\big),
\label{eq:mv}
\end{equation}
where $\lambda=0$ results in a risk-neutral policy and high-variance actions are penalised for $\lambda>0$. Since the mean--variance objective does not satisfy the principle of optimality \citep{sobel1982variance,mannor2011meanvariance}, the criterion is applied during decision-making to the learned distribution, and the resulting policies are empirically characterised in Section~\ref{section:results}. Bootstrap actions in the distributional Bellman target are selected with the same risk-adjusted score and the same mask.

\section{Experiments and Results}
\label{section:results}

\subsection{Experimental setup}
\label{subsec:setup}

The framework is demonstrated on five-target campaigns inspired by the Iridium-33 debris cloud \citep{kelso_celestrak}. This catalogued debris cloud extends from roughly 405 to \SI{900}{km} in altitude, with inclinations near $86.4^\circ$ and almost circular orbits. It is a densely populated, congested LEO region whose members share similar inclinations. Differential nodal precession has since spread the ascending nodes of the fragments, and the targets in the scenario are taken to share a common node, for which transfer costs depend on altitude and small inclination changes. Targets are chosen with altitudes uniformly distributed between 500 and \SI{900}{km} per episode, and inclinations within $86.4^\circ\pm0.65^\circ$.

\begin{figure}[t]
\centering
\includegraphics[width=0.94\columnwidth]{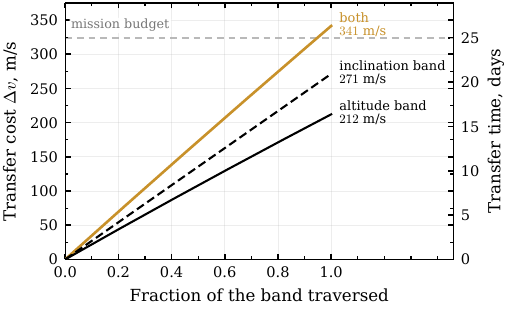}
\caption{Cost of traversing the sampled orbital bands, starting from the orbit at \SI{500}{km} and $85.75^\circ$, calculated using Eq.~(\ref{eq:edelbaum}). The right axis indicates the equivalent thrusting time. A transfer across both bands exceeds the mission budget.}
\label{fig:tcost}
\end{figure}

According to Eq.~(\ref{eq:edelbaum}), crossing the entire altitude range requires a $\Delta v$ of \SI{212}{m/s}, while crossing the inclination range requires a $\Delta v$ of \SI{271}{m/s}, meaning transfer costs depend on both parameters. Figure~\ref{fig:tcost} illustrates the costs for each. We consider a servicer with a thrust of \SI{0.75}{N} and a mass of \SI{5000}{kg}, which gives an acceleration of $1.5\times10^{-4}$\,\si{m/s^2}, of the same order as the low-thrust servicer of \citet{apa_optimal_2025-1} (\SI{0.5}{N} and \SI{2000}{kg}). The velocity increment is therefore \SI{12.96}{m/s} per day. For a 25-day mission, the budget is \SI{324}{m/s}. Crossing both ranges costs \SI{341}{m/s}, exceeding the mission budget, so it is not possible to reach both of the most widely separated targets within the same mission.

The servicer begins at a random target, which is considered inspected, and may perform up to four inspections per episode. Inspection values are drawn uniformly between $1$ and $4$. The risk state is redrawn every three decisions, with a premium $\beta=2.5$, a failure probability $p=0.15$, and a loss $L=8$. For example, a high-risk inspection of a target with value $2.5$ has an expected increase of $0.55$ over its nominal value, with a possibility of a loss of $L$. Table~\ref{tab:params} summarises these parameters.

\begin{table}[t]
\centering
\caption{Environment and training parameters for the five-target scenario.}
\label{tab:params}
\footnotesize
\begin{tabular}{@{}lll@{}}
\toprule
Symbol & Description & Value \\
\midrule
$N$ & candidate targets & 5 \\
    & target altitude range & \SIrange{500}{900}{km} \\
    & target inclination range & $86.4^\circ\pm0.65^\circ$ \\
$F,\,m$ & thrust, servicer mass & \SI{0.75}{N},\ \SI{5000}{kg} \\
$B$ & mission time budget & \SI{25}{days} \\
$v_j$ & inspection value & $\mathcal{U}[1,4]$ \\
$\beta$ & high-risk success premium & 2.5 \\
$L$ & high-risk failure loss & 8 \\
$p$ & failure probability & 0.15 \\
$n_e$ & high-risk targets / update period & 2 / every 3 steps \\
$K$ & quantiles per action & 200 \\
    & network (hidden layers) & $256\times256$ \\
    & learning rate / exploration fraction & $10^{-4}$ / 0.5 \\
$\gamma$ & discount factor (training only) & 0.99 \\
    & training steps / seeds & $1.5\times10^{6}$ / 3 \\
    & evaluation episodes & 4000 \\
$\lambda$ & mean--variance weights & $\{0,\ 0.1,\ 0.3\}$ \\
\bottomrule
\end{tabular}
\end{table}

A new instance is sampled at the start of each episode, so policies are trained and tested across the scenario distribution. For each agent configuration ($\lambda=0$, $0.1$, $0.3$), three random seeds are used for training, and evaluation is conducted with a deterministic policy on the same 4000 test instances for all policies, sampled from an independent seed stream. Results are averaged across these seeds. The enumeration of Section~\ref{subsec:exact} is run on the same instances, with between $4$ and $24$ admissible visit orders per instance and $12.5$ on average. A baseline policy that selects uniformly at random among the admissible actions at each step is also evaluated on the same instances, averaged over five random seeds. This baseline uses the same feasibility mask and continues to inspect as long as any uninspected target remains affordable, differing from the learned policies only in the action selected. Performance is summarised using the mean mission return $Z$, its variance, the lower decile $q_{0.1}$, the probability of a mission loss ($Z<0$), and the number of targets inspected.

\subsection{Exact efficient frontier}
\label{subsec:exact}

The efficient frontier is the set of policies for which no other admissible policy attains a higher mean return with an equal or lower return variance. It gives, for each level of return variance, the highest mean return achievable on the evaluation instances, and it serves here as the reference against which the learned policies are measured. At the parameters of Table~\ref{tab:params}, it can be computed exactly. The inspection outcome affects only the collected reward. The position of the servicer, the remaining budget and the set of uninspected targets depend only on the targets chosen (Section~\ref{subsec:mdp}). The risk state is redrawn once within an episode, after the third inspection, when a single target remains uninspected, and that target is then in the high-risk state. The sequence of observations is therefore determined by the visit order, and among policies that act on the observation, an optimal policy is a predetermined visit order. A visit order is admissible if each transfer fits within the remaining budget and no uninspected target is affordable at the end of the sequence, consistent with the campaign rule of Section~\ref{subsec:mdp}. Inspection outcomes are independent along the sequence, so the conditional mean and variance of $Z$ are sums of the per-inspection moments of Eq.~(\ref{eq:demand}). The total variance over the evaluation set is the sum of the within-instance variance and the variance of the conditional mean across instances. Order selection is optimised based on this pooled measure using fixed-point iteration. Varying $\lambda$ across admissible orders traces the frontier of mean return versus return variance for the scenario distribution. Evaluated policies are then compared to this frontier at their achieved variance level, through interpolation between known values.

\subsection{Aggregate performance}
\label{subsec:aggregate}

Table~\ref{tab:results} reports the evaluation of the three learned configurations, of uniform selection and of the optimum. The optimum, obtained by enumeration at $\lambda=0$, is a mean return of $9.60$ on these instances. The learned risk-neutral policy reaches $8.96\pm0.04$, or $93.3\%$ of the optimum, and inspects $2.99$ of the four remaining targets per campaign. Uniform selection among the admissible targets reaches $73.0\%$ of the optimum, and the worst admissible order on each instance yields a mean return of $4.43$.

The value collected by a campaign therefore depends strongly on which targets are inspected and in which order, since, averaged over the instances, the mean return ranges from $4.43$ for the worst admissible order to $9.60$ for the best. The learned policy recovers three quarters of the difference between uniform selection and the optimum, with each decision taken from the current observation among at most $N$ admissible targets. For mission planning, this places the closed-loop planner within $7\%$ of the best campaign obtainable by full enumeration, without relying on an enumeration that grows combinatorially with the number of targets.

\begin{table}[t]
\centering
\caption{Performance of each policy on the same 4000 test instances, $N=5$. Values for the learned policies are means over three training seeds, with standard deviations no more than $0.05$ for the mean of $Z$ and $0.42$ for the variance of $Z$. Uniform selection is averaged over five randomisation seeds, with standard deviations of $0.06$ and $0.40$.}
\label{tab:results}
\footnotesize
\setlength{\tabcolsep}{3.6pt}
\begin{tabular}{@{}lS[table-format=1.2]cS[table-format=-1.2]cc@{}}
\toprule
Policy & {$\mathbb{E}[Z]$} & {$\mathbb{V}[Z]$} & {$q_{0.1}$} & {$P(Z{<}0)$} & {Inspected} \\
\midrule
uniform selection       & 7.01 & 34.0 & -1.99 & 13.1\% & 2.47 \\
\midrule
QR-DQN, risk-neutral    & 8.96 & 42.3 & -1.21 & 12.9\% & 2.99 \\
QR-DQN, $\lambda=0.1$   & 7.84 & 28.7 & 0.02 & 10.0\% & 2.77 \\
QR-DQN, $\lambda=0.3$   & 6.66 & 20.2 & 2.42 & \phantom{0}7.5\% & 2.52 \\
\midrule
optimum & 9.60 & 49.9 & \multicolumn{3}{l}{exact, by enumeration} \\
\bottomrule
\end{tabular}
\end{table}

\subsection{The risk--return trade-off}
\label{subsec:frontier}

\begin{figure}[t]
\centering
\includegraphics[width=0.94\columnwidth]{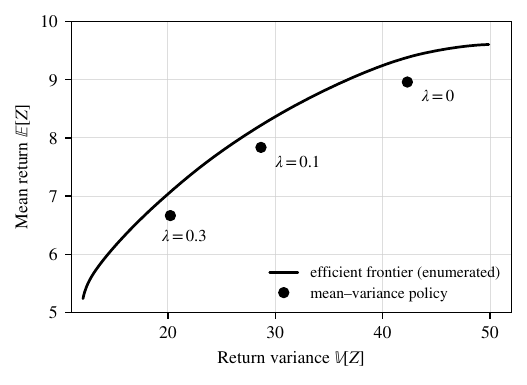}
\caption{Mean return against return variance on the same 4000 test instances. The solid line is the efficient frontier, obtained by enumeration. Filled circles are the mean--variance policies for $\lambda=0$, $0.1$ and $0.3$, with error bars showing the standard deviation across three training seeds.}
\label{fig:pareto}
\end{figure}

Figure~\ref{fig:pareto} plots the mean return of each policy against its return variance, and Table~\ref{tab:results} lists the lower decile and the loss probability. As $\lambda$ increases, the mean return and the return variance both decrease. From $\lambda=0$ to $\lambda=0.3$, the return variance falls from $42.3$ to $20.2$, about one half, the lower decile rises from $-1.21$ to $+2.42$, and the mean return decreases from $8.96$ to $6.66$.

Figure~\ref{fig:pareto} also shows the efficient frontier. At the return variances of the learned policies, $42.3$, $28.7$ and $20.2$, the frontier mean returns are $9.38$, $8.22$ and $7.07$, against $8.96$, $7.84$ and $6.66$ for the learned policies, which therefore attain $94\%$ to $96\%$ of the highest mean return available at their variance. The lower tails differ. At the frontier points of nearest variance ($28.3$ and $20.5$), the lower decile is $-2.20$ and $-2.49$ and the loss probability is $14.2\%$ and $11.6\%$. The learned policies with $\lambda=0.1$ and $0.3$ reach a lower decile of $0.02$ and $2.42$ and a loss probability of $10.0\%$ and $7.5\%$.

The weight $\lambda$ thus acts as a single control on the risk--return balance of the campaign, with the environment and the reward unchanged, and the learned policies remain close to efficient across its range. A weight can be selected for a required loss probability or lower decile, and the mean return given up at that tolerance can be read from Table~\ref{tab:results}. At equal variance, the reduction obtained by the risk-averse policies is concentrated in the loss tail. This is consistent with the two variance measures involved. The frontier minimises the variance pooled over all instances, which includes the spread of the conditional mean return between instances. The learned score, applied at each decision (Section~\ref{subsec:mv}), penalises the variance of the return predicted from the current state, which is dominated by the outcomes of high-risk inspections. A few percent of mean return at a given variance are thereby exchanged for a markedly lower probability of a mission loss, the quantity of primary concern in a safety-critical mission.

\subsection{Return distributions}
\label{subsec:distributions}

\begin{figure}[t]
\centering
\includegraphics[width=0.91\columnwidth]{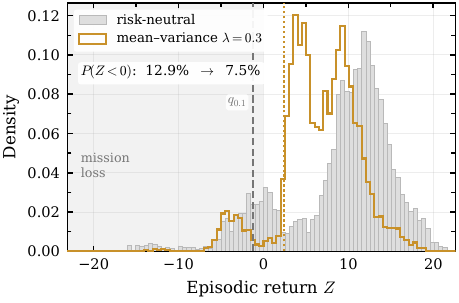}
\caption{Return distributions for the risk-neutral and mean--variance ($\lambda=0.3$) policies, combined across three training seeds, with $12\,000$ evaluations each. Dashed and dotted verticals mark the lower decile ($q_{0.1}$) of each distribution. The shaded region marks mission losses ($Z<0$).}
\label{fig:distributions}
\end{figure}

Figure~\ref{fig:distributions} shows the return distributions of the risk-neutral and $\lambda=0.3$ policies, pooled over three training seeds. The risk-neutral distribution has more mass at high returns and a heavier loss tail, and the $\lambda=0.3$ distribution has more mass at moderate returns. The probability of a mission loss, $P(Z<0)$, decreases from $12.9\%$ to $7.5\%$, and the worst return over the $12\,000$ evaluations improves from $-22.6$ to $-14.8$. The reduction in mean return comes mainly from the upper part of the distribution.

The mean--variance criterion penalises deviations on both sides of the mean. In this problem its effect falls mainly on the lower tail, because negative returns arise only from failed high-risk inspections. Risk aversion therefore reduces the number of mission losses and limits the worst outcome, and the mean return given up comes mainly from the upper part of the distribution, where the high returns of successful high-risk inspections lie. For the operator, giving up about a quarter of the expected value at $\lambda=0.3$ reduces the probability of a mission loss by about $40\%$.

\section{Conclusion}
\label{section:conclusions}

Multi-target servicing campaigns are planned before the outcome of each operation is known, and a failed operation can cost more than several successful ones. The operator must then decide how much expected value to give up for protection against loss. This paper formulates the planning of such campaigns as a risk-sensitive MDP. The formulation includes a mission-time budget, closed-form low-thrust transfer costs and a stochastic risk of failure at each target. In the proposed distributional RL planner, the operator's decision is expressed through a single mean--variance weight applied to the learned return distribution. The planner was demonstrated on five-target inspection campaigns inspired by the Iridium-33 debris cloud. This size was chosen so that the efficient frontier of mean return against return variance could be computed exactly by enumerating the admissible visit orders and used to validate the learned policies. The learned policies attain $94$ to $96\%$ of the frontier mean return at matched variance. With $\lambda=0.3$, the return variance is halved and the probability of a mission loss falls from $12.9\%$ to $7.5\%$, at the cost of about a quarter of the mean return. At equal variance, the learned risk-averse policies also have a lower loss probability than the frontier policies. The weight can therefore be set from a required loss probability or lower decile, with the mean return given up at that tolerance known from evaluation.

For larger campaigns, the number of admissible orders grows combinatorially, and an exact reference of this kind is no longer available. The learned planner selects among at most $N$ actions at each step, so the cost of each decision grows only linearly with the number of targets, and the planner is therefore scalable. The orbital and risk models are simplified, and more detailed models would alter the transfer costs and inspection risks. However, the core formulation remains unchanged, since a different transfer law or risk model affects only the leg costs, the risk-state updates and the inspection outcome distributions.

\section*{Acknowledgments}
This work was supported by the \emph{Agence Nationale de la Recherche (ANR)}, France, under the project \textbf{\mbox{ANR-23-CE10-0006}} (Resilient and Sustainable Planning and Management of Future Space Industry Infrastructure with On-Orbit Servicing).

\addcontentsline{toc}{section}{References}
\bibliographystyle{unsrtnat}
{
\renewcommand{\refname}{\normalsize References}
\setlength{\bibsep}{1.5pt plus 0.3ex}
\footnotesize \bibliography{references}
}

\end{document}